\documentclass[letterpaper]{article} 
\usepackage{aaai25}  
\usepackage{times}  
\usepackage{helvet}  
\usepackage{courier}  
\usepackage[hyphens]{url}  
\usepackage{graphicx} 
\usepackage{soul}
\usepackage{natbib}  
\usepackage{caption} 
\usepackage{algorithm}
\usepackage[switch]{lineno}

\usepackage[utf8]{inputenc} 
\usepackage{url,enumitem}            
\usepackage{booktabs}       
\usepackage{amsfonts}       
\usepackage{nicefrac}       
\usepackage{microtype}      
\usepackage{xcolor}         
\usepackage{newfloat}
\usepackage{listings}
\DeclareCaptionStyle{ruled}{labelfont=normalfont,labelsep=colon,strut=off} 
\floatstyle{ruled}
\newfloat{listing}{tb}{lst}{}
\floatname{listing}{Listing}
\usepackage{amsmath,amsthm,amssymb}
\usepackage{graphicx}
\usepackage{bm}
\usepackage{subfig}
\usepackage{array}
\usepackage{multirow}
\usepackage{makecell}
\usepackage{color, colortbl}
\usepackage{cleveref}
\usepackage{enumitem}
\usepackage{pifont}
\usepackage{algorithm}
\usepackage{algorithmic}

\usepackage{tikz}
\usepackage[symbol]{footmisc}
\newcommand*\circled[1]{\tikz[baseline=(char.base)]{ \node[shape=circle,draw,inner sep=0.2pt] (char) {#1};}}

\crefname{section}{Sec.}{Secs.}
\Crefname{section}{Section}{Sections}
\Crefname{table}{Table}{Tables}
\crefname{table}{Tab.}{Tabs.}
\graphicspath{{./figs/}}
\title{
    FlexPose: Pose Distribution Adaptation with Limited Guidance
}

\iftrue
\author {
    Zixiao Wang\textsuperscript{\rm 1}~~
    Junwu Weng\textsuperscript{\rm 2}\footnote{Corresponding Author}~~
    Mengyuan Liu\textsuperscript{\rm 3}~~
    Bei Yu\textsuperscript{\rm 1}\footnotemark[1]
}
\affiliations {
    \textsuperscript{\rm 1}The Chinese University of Hong Kong~~~\textsuperscript{\rm 2}ByteDance Inc.~~~\textsuperscript{\rm 3}Peking University\\
    zxwang22@cse.cuhk.edu.hk,~ we0001wu@e.ntu.edu.sg,~ 
    nkliuyifang@gmail.com,~
    byu@cse.cuhk.edu.hk
}
\fi

\usepackage{bibentry}
\usepackage{comment}

\newcommand{\minisection}[1]{\noindent{\textbf{#1}.}}

\begin{document}

\maketitle

\begin{abstract}

Numerous well-annotated human key-point datasets are publicly available to date. However, annotating human poses for newly collected images is still a costly and time-consuming progress. Pose distributions from different datasets share similar pose hinge-structure priors with different geometric transformations, such as pivot orientation, joint rotation, and bone length ratio. 
The difference between Pose distributions is essentially the difference between the transformation distributions. Inspired by this fact, we propose a method to calibrate a pre-trained pose generator in which the pose prior has already been learned to an adapted one following a new pose distribution. We treat the representation of human pose joint coordinates as skeleton image and transfer a pre-trained pose annotation generator with only a few annotation guidance. By fine-tuning a limited number of linear layers that closely related to the pose transformation, the adapted generator is able to produce any number of pose annotations that are similar to the target poses. 
We evaluate our proposed method, FlexPose, on several cross-dataset settings both qualitatively and quantitatively, which demonstrates that our approach achieves state-of-the-art performance compared to the existing generative-model-based transfer learning methods when given limited annotation guidance.

\end{abstract}

\section{Introduction}

Deep neural networks are data-hungry and rely on large-scale datasets with high-quality human annotations for training. However, the process of annotating these datasets can be expensive and time-consuming, particularly when dense annotation are required, as is often the case in pose estimation tasks~\cite{wang2022contextual,he2017mask}. To overcome this challenge, AI-aided labeling methods have become increasingly popular, where a pre-trained model's prediction serves as a reference to reduce human workload. However, when there is a domain shift~\cite{luo2019taking}, where the distribution of the training dataset and test dataset are not aligned not only on the input image domain but on the pose annotation domain as well, the accuracy of the model can significantly decline.

Considerable efforts have been devoted to tackling this issue.
Among them, domain adaptation (DA)~\cite{daume2009frustratingly,csurka2017domain} introduces knowledge from existing annotated datasets to a target dataset and is verified effective on several computer vision tasks~\cite{cao2019cross,inoue2018cross}.
However, things become different in the human-related dataset, {\it e.g.}, human pose~\cite{h36m_pami} and human face~\cite{wayne2018lab}.
As the source human appearance is usually required in DA-related methods for input image domain adaptation,
they may import unexpected data distribution bias~\cite{buolamwini2018gender}, {\it e.g.}, gender or color, from the source.
Besides, the direct exposure of private portraits may raise the privacy issue.

\begin{figure}
    \centering
    \includegraphics[width=.93\linewidth]{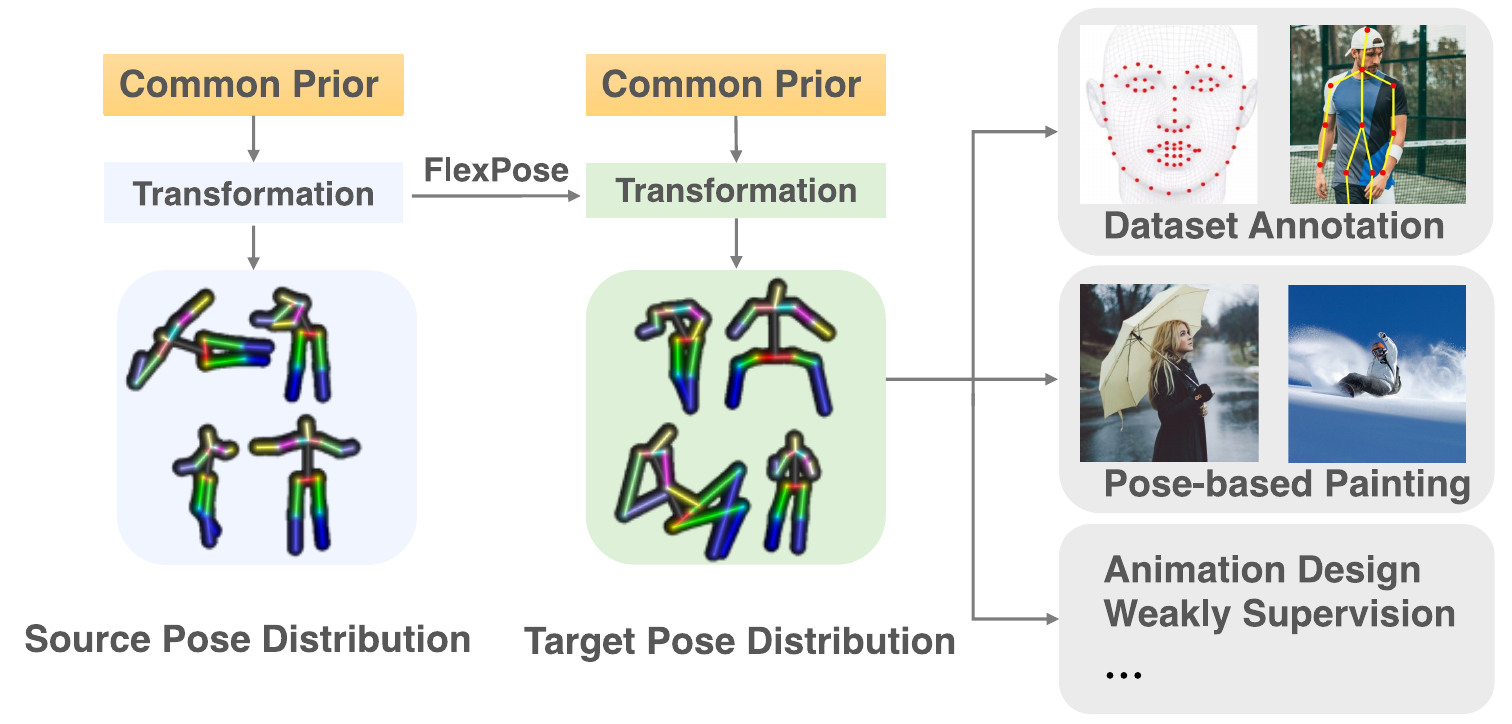}
    \caption{The illustration of how poses can be adapted between different domains. Although various pose datasets may differ in their transformations, they share a common hinge-structure prior. FlexPose's adaptation process focused on transformation, and the resulting poses can be effectively used in a wide range of downstream pose-related tasks.}
    \label{Fig:diffdata}
\end{figure}

On the other hand, it is commonly observed that different human poses share a similar hinge-structure prior. Typically, poses in a target dataset can be transferred from poses of a pre-collected source set by applying geometric transformations on for example pivot orientation, joint rotation, and bone length ratio. Therefore, adapting the pose distribution {\it only} can be a viable option in the case of human-related datasets. Pose Domain Adaptation (PDA) avoids the direct use of human appearance images, effectively addressing the aforementioned issues. Motivated by this observation, we propose FlexPose (shown in \Cref{Fig:diffdata}), a method that transfers the source pose distribution to a target distribution with limited pose annotation guidance. After pose distribution transfer by FlexPose, each {\it input image} can be matched with the most related {\it generated pose} in estimated pose distribution by utilizing a matching algorithm~\cite{jakab2020self} for weakly supervised pose estimation and pose annotation. Besides, the generated poses can also be utilized in plenty of downstream tasks such as pose-conditioned image generation~\cite{zhang2023adding}. 

In FlexPose, we treat pose annotations as {\it skeleton images} to well align the annotations with their RGB appearance correspondences,
and to improve the learnability of pose prior as the skeleton images well preserve the spatial structure of joint connection on image plane.
We first learn the pose prior and fit the empirical distribution from a source human pose dataset by a multi-layer generative model.
Thereafter, specific layers of the generative model are calibrated by inserting learnable lightweight linear modules to transfer the source distribution to the target domain.
Considering that only a limited number of poses are given, we introduce three regularizations to avoid the collapse of the transfer solution.
By generating credible pose interpolations with {\it Pose-mixup} regularization and by strictly limiting the complexity of the transfer module with linear and sparse regularization,
we minimize the requirement of sample amount but maximize the sample diversity in FlexPose.
FlexPose is computation-efficient. It operates on the pose domain, and hence the training convergence is much faster than methods in domain adaptation,
which focuses adaptation on both the pose and image domains together.
FlexPose is also data-efficient. We only need limited pose annotations from the target dataset to finetune the transfer modules. 
Extensive experiments on three pose-based tasks, {\it i.e.,} human pose annotation, human face landmarks annotation, and pose-conditional human image generation,
demonstrate that FlexPose outperforms baselines by a large margin both quantitatively and qualitatively.
Our contributions can be summarized as follows:
\begin{itemize}
    \item We propose to treat the task of Pose Domain Adaptation as the transfer of skeleton image generator and demonstrate that a target pose distribution can be well approximated from a well-learned pose prior.
    \item We introduce FlexPose, a PDA framework that employs three regularizations to efficiently transfer a pose distribution to a target one by utilizing a limited number of guiding poses with low computation and storage costs.
    \item Extensive experimental results on three pose-related tasks show that FlexPose achieves remarkable improvement over existing methods. 
\end{itemize}

\section{Related Works}

\begin{figure*}[tb!]
    \centering
    \includegraphics[width=.88\textwidth]{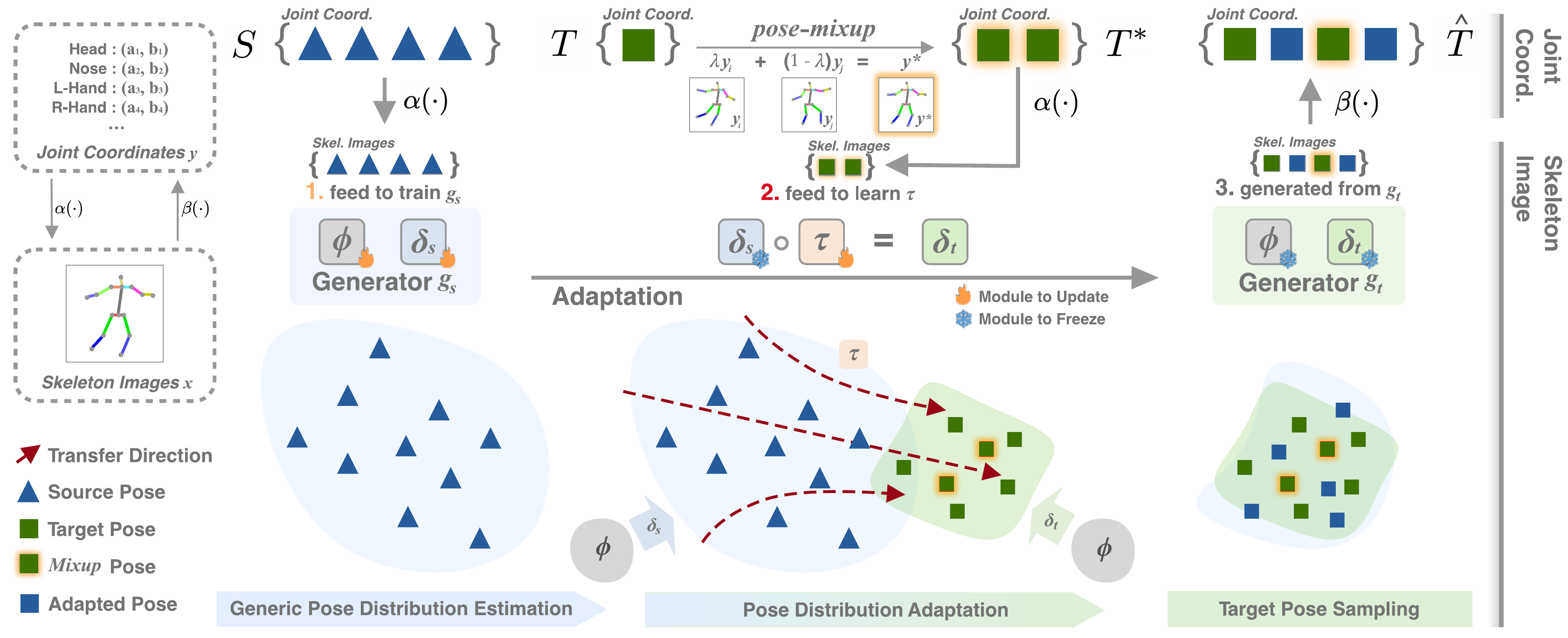}
    \caption{
        An illustration of the FlexPose framework for pose distribution adaptation.
        There are three main steps in our framework:
        \circled{1}
        We train a skeleton image generator to learn the pose prior from the source pose distribution; 
        \circled{2}
        The source generator is transferred to a target generator with limited target pose guidance to achieve pose distribution adaptation;
        \circled{3}
        We utilize the target generator to generate target pose annotations for downstream tasks.
    } 
    \label{fig:overview}
\end{figure*}

\minisection{Deep Generative Model for Image Generation}
Deep generative models such as GAN, Variational AutoEncoder (VAE), and Diffusion models achieve great success in realistic/artificial image generating and natural image distribution modeling.
Recently proposed generative models such as StyleGAN~\cite{karras2019style}, DDPM~\cite{ho2020denoising}, NVAE~\cite{vahdat2020nvae} introduce new mechanisms, new architecture, and new regularizations into image generation.
VAEs~\cite{kingma2013auto} learn to maximize the variational lower bound of likelihood. Diffusion probabilistic models~\cite{sohl2015deep} synthesize images by a denoising procedure.
GANs~\cite{goodfellow2014generative} are trained in an adversarial manner to learn how to generate realistic images.
Among them, Karras {\it et al.}~\cite{karras2019style} proposed an architecture StyleGAN that can learn a hierarchical decoupled style code and controls image synthesis. 
Our method is based on generators with multi-layer architecture and leverages StyleGAN as the backbone.
We are inspired by the recent works~\cite{zhu2016generative,yin2022styleheat}, which manipulate the latent code in the generative model to edit the output images. These works motivate us to transfer the pose distribution to the target domain by transferring style codes with few-shot guidance.

\minisection{Transfer Learning for Generative Models}
The literature on transfer learning has been extensively studied in recent years \cite{oquab2014learning, long2015learning,ganin2015unsupervised}.
Transfer learning learns to transfer the knowledge from a large-scale source dataset to a small target dataset to enhance model performance on the target dataset.
The methodology of transfer learning is also treated as a pre-training technique.
It is utilized to accelerate the learning on the target dataset.
\cite{wang2018transferring} finetunes a pre-trained GAN on a target dataset to get better performance.
\cite{noguchi2019image} transfers knowledge from a large dataset to a small dataset by re-computing batch statistics.
Existing methods focus on either the image domain or the neural language processing domain \cite{shin2016generative}.
For these methods, hundreds of training samples are still required.
Compared with these approaches, we focus on pose domain adaptation, and our method only requires few-shot guidance for transferring.
After LoRA~\cite{hu2021lora} is widely used in Large Language Model finetuning, the researchers in Content Generation are inspired to introduce or extend this technique in generative model~\cite{mou2024t2i}.
Compared with the light-weight but global model finetuning in LoRA, FlexPose only focuses on calibrating specific layers with semantics of pose geometric transformation locally to satisfy the linear and sparse finetuning requirements.

\minisection{Human Pose Estimation}
2D Human pose estimation is a task that predicts the 2D pose from a single image.
Fully-supervised methods~\cite{andriluka2009pictorial,bai2019acpnet,belagiannis2017recurrent} utilize large-scale annotated datasets such as COCO~\cite{lin2014microsoft}, Human3.6M~\cite{h36m_pami} and 3DHP~\cite{mehta2017monocular} for model training.
Weakly-supervised~\cite{kanazawa2018end,gecer2019ganfit,geng20193d,wang2023truncate} and unsupervised~\cite{shu2018deforming,jakab2018unsupervised} methods such as KeypointGAN~\cite{jakab2020self} have been proposed to reduce the dependence on the expensive pose annotation. 
These methods require supervised post-training or additional prior knowledge to generate meaningful landmarks, which can serve as a distance measurement between the provided prior knowledge and the target distribution. To match poses generated by FlexPose with unlabelled images in the target dataset, we employ an unsupervised method~\cite{jakab2020self} in addition to supervision from adversarial training. This matching procedure serves as an evaluation method for FlexPose and is further detailed in \Cref{sec:4}. 
Recently, test-time adaptation~\cite{li2021test,cui2023test,hu2024fast} has proven to be an effective way to deal with domain shift in pose estimation. It utilizes self-supervised learning during inference to adapt model to the input human appearance distribution. Compared with FlexPose which focuses on PDA, Test-time adaptation tackle the issues in input image domain shift.

\section{Method}

Given a limited number of 2D pose annotations set ${T}=\left\{\bm{y}_t|\bm{y}_t\in \mathbb{R}^{M\times 2}\right\}$ of a newly collected human pose images, FlexPose aims to estimate the whole distribution $\mathcal{D}_t$ which pose annotation $T$ belongs to, and generate any number of new pose annotations that follow the distribution. $M$ here is the number of joints in each pose. This task setting is challenging. However, we believe that with the prior from sufficient off-the-shelf annotations $S = \left\{\bm{y}_s|\bm{y}_s\in \mathbb{R}^{M\times 2}\right\}$, the distribution $\mathcal{D}_t$ can be estimated and well shaped. In this paper, we transfer the distribution $\mathcal{D}_s$ estimated from $S$ to the target pose domain to estimate the target distribution $\mathcal{D}_t$ by considering the guidance from 2D pose annotations set $T$.

\subsection{Overview}

As illustrated in \Cref{fig:overview}, our framework consists of three phases:
\circled{1} 
\textbf{Generic Pose Distribution Estimation}.
We learn a generator $g_s(\cdot)$ on the pose set $S$ to estimate the pose distribution $\mathcal{D}_s$.
The generator takes a latent code $\bm{z}$ as input and outputs a skeleton $\hat{\bm{x}}_s$, {\it i.e.} $\hat{\bm{x}}_s=g_s(\bm{z})$.
We take the distribution of generated $\hat{\bm{x}}_s$ to mimic that of the generic pose $\bm{x}_s$.
Here, $\bm{x}$ is the corresponding {\it skeleton image} of an annotation $\bm{y}$ as shown in the left part of \Cref{fig:overview}.
\circled{2} 
\textbf{Pose Distribution Adaptation}.
Given the limited target annotation set $T$, we transfer $g_s(\cdot)$ to fit the pose distribution $\mathcal{D}_t$ and learn a new generator $g_t(\cdot)$ of the target pose domain.
Considering the limited knowledge acquired from target pose annotation $T$, we introduce three regularizations, {\it Linear, Sparse} and {\it Pose-mixup}, to avoid reaching a collapse solution.
\circled{3} 
\textbf{Target Pose Sampling}.
The transferred generator $g_t(\cdot)$ can flexibly synthesize any number of fake pose annotations by randomly sampling in the latent space.
This generated annotation set $\hat{T}$ will be treated as an extension of given annotations set $T$ in the downstream tasks, {\it e.g.,} Keypoints Annotation and Pose-conditional Human Image Generation,
since poses within both of them follow the distribution $\mathcal{D}_t$.

\subsection{Generic Pose Distribution Estimation}
\label{sec:3.2}

Deep generative models have been widely verified that they have a rich capacity to well approximate image distributions when given sufficient training data. Motivated by the success of these generative models~\cite{karras2019style} on natural/artificial image generation, we treat 2D pose annotations $\bm{y}_s, \bm{y}_t\in \mathbb{R}^{M\times 2}$ as skeleton images $\bm{x}_s, \bm{x}_t \in \mathbb{R}^{C\times W \times H}$ and extend an image generator to generate 2D pose annotations by synthesizing corresponding skeleton images. As shown in the left part of \Cref{fig:overview}, the transformation from the 2D keypoints to the skeleton images can be implemented by functions $\alpha(\cdot)$, namely $\bm{x}=\alpha(\bm{y})$, where $\alpha(\cdot)$ simply draws keypoints from $\bm{y}$ and connects them with straight lines on a blank figure.
The visual effect is similar to the stick man. To achieve precise semantic alignment with the appearance correspondence, each bone in the skeleton image is assigned a unique color. Therefore, $C$ of each skeleton image is set as three (RGB channels). Compared with Black\&White, the colorful embedding brings marginal improvement in the quality of generated skeletons.

A generator can be formulated as a mapping function $g(\cdot)$, which gets a latent code $\bm{z}$ and outputs a skeleton image $\bm{x}$. The probability distribution of skeleton images hence is estimated by $p(\bm{x})=p(\bm{z})p_g(\bm{x}|\bm{z})$. We assume that the pose distributions of different datasets share similar pose prior, and their distributions can transfer to one another by geometric transformations. Based on this assumption, we further factorize the generator $g(\cdot)$ as $g=\phi\circ\delta $. Therefore, the source skeleton image generator can be formulated as 
\begin{linenomath}
\begin{equation}
    p(\hat{\bm{x}}_{s}) = p(\bm{z})~p^s_g(\hat{\bm{x}}_{s}|\bm{z}) = p(\bm{z})~p_{\delta}^s(\bm{h}_s|\bm{z})~p_{\phi}(\hat{\bm{x}}_{s}|\bm{h}_s),
    \label{eq:src_gen}
\end{equation}
\end{linenomath}
in which $\phi(\cdot)$ preserves the learned pose prior and $\delta(\cdot)$ records the mapping from the learned prior $\bm{h}$ to the skeleton image $\bm{x}_s$ of a certain pose domain.
Similarly, the distribution of target domain can be formulated as $p(\hat{\bm{x}}_t)=p(\bm{z})~p_{\delta}^t(\bm{h}_t|\bm{z})~p_{\phi}(\hat{\bm{x}}_t|\bm{h}_t)$.
With the prior sharing assumption, the pose distribution adaptation aims at transferring pre-trained conditional probability $p_{\delta}^s(\bm{h}_s|\bm{z})$ to $p_{\delta}^t(\bm{h}_t|\bm{z})$ with guidance from the pose annotation set $T$:
\begin{linenomath}
\begin{equation}
    p_{\delta}^s(\bm{h}_s|\bm{z})\stackrel{T}{\longrightarrow}p_{\delta}^t(\bm{h}_t|\bm{z}).
    \label{eq:guide}
\end{equation}
\end{linenomath}

Considering the ability of StyleGAN in separating high-level attributes and in the interpolation between these attributes,
we utilize StyleGAN network architecture to disentangle the pose prior and the transformation of the source skeleton image generator,
\begin{linenomath}
\begin{equation}
\label{eq:stylegan}
    g_s = \phi \circ {\delta}_s = \phi \circ (A \circ f)_s,
\end{equation}
\end{linenomath}
where $f(\cdot)$ is a non-linear mapping that takes random noise as input and outputs a random vector.
$A(\cdot)$ is a learned affine transformation and can be treated as a block diagonal matrix with $L$ blocks, where $L$ is the number of layers.
The output of $A(\cdot)$ is the style code to modulate the synthesis network $\phi(\cdot)$ by adaptive instance normalization.
Due to the ability of StyleGAN in style control, we can directly adapt the distribution of source skeleton image to the target domain by adjusting the style code.

\begin{figure}[tb!]
    \centering
    \includegraphics[width=0.8\linewidth]{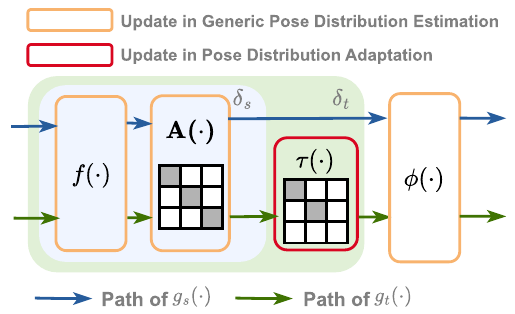}
    \caption{An illustration of the generator decomposition. We use $\tau(\cdot)$ to adjust the source generator for pose distribution adaptation.}
    \label{fig:network}
\end{figure}

\subsection{Pose Distribution Adaptation}
\label{sec:3.3}

As illustrated in \Cref{fig:network}, to transfer $p_{\delta}^s(\bm{h}_s|\bm{z})$ to $p_{\delta}^t(\bm{h}_t|\bm{z})$, we adjust the style code by introducing a transfer function $\tau(\cdot)$ at the output of $\delta(\cdot)$, and therefore the target domain generator is defined as
\begin{linenomath}
\begin{equation}
    g_t = \phi \circ {\delta}_t = \phi \circ (\tau \circ {\delta}_s).
\label{eq:h}
\end{equation}
\end{linenomath}

To learn the transfer function $\tau$, we first randomly sample $|T|$ latent codes $\bm{z}$ (one for each pose in $T$) from the latent space, and require the generator maps each code to corresponding skeletons in $T$. This transferring procedure can be achieved by minimizing the following perceptual loss,
\begin{linenomath}
\begin{equation}
\label{eq:rec}
    \min_{\theta_\tau}\mathcal{L}_{s\rightarrow t} = \min_{\theta_\tau}\sum \Big\Vert\Gamma\big(g_t(\bm{z});\theta_{\tau}\big) - \Gamma\big(\bm{x}\big)\Big\Vert^2_2,
\end{equation}
\end{linenomath}
where $\theta_\tau$ is the parameter of $\tau(\cdot)$, $\Gamma$ is a pre-trained feature extractor, $\bm{z}$ is from the set of sampled latent codes, and $\bm{x}$ is the skeleton image drawn from the pose annotation set $T$. 

However, the problem is we only have few-shot guidance $T$ from the target domain distribution. Given a data-starving deep learning model, the guidance is insufficient to reach a satisfactory solution. For that reason, we introduce three regularizations to alleviate the data-insufficient issue.

\noindent\textbf{Linear $\&$ Sparse Regularization}.
Compared with finetuning the whole transformation function ${\delta}_s$ to reach ${\delta}_t$, only adjusting the affine transformation from $A_s$ to $A_t$, {\it i.e.} $A_t = \tau \circ A_s$,  can efficiently shrink the searching space of transfer solution, and therefore avoid overfitting. 
Meanwhile, the recent GAN inversion technique shows that the layer-wise style code in StyleGAN leads to the hierarchical disentanglement of local and global attributes, which aligns well with our motivation of adapting pose distribution by considering the global geometric transformation between poses. We thus adjust the source affine transformation $A_s$ from the perspective of layer level, and limit the number of to-be-adjusted layers as small as possible. Considering the form of the affine transformation $A$ and the layer decoupling characteristics of StyleGAN, we empirically define the transfer function $\tau(\cdot)$ as a block diagonal matrix, 
\begin{linenomath}
\begin{equation}
    \tau\triangleq \text{diag}(\bm{I},...,\bm{I}, \bm{U}_{l_0}, \bm{I},...,\bm{I}, \bm{U}_{l_1}, \bm{I},...,\bm{I}),
\end{equation}
\end{linenomath}
where only a limited number of block is defined by $\bm{U}$, {\it i.e.} $l_0$ and $l_1$ in this case, to follow the sparse regularization. We experimentally find that the earlier layers are most related to the geometric transformation. And we only learn those layers in our experiments. Meanwhile, other blocks are set as identity matrix $\bm{I}$. We investigated how the choice of layer $l$ affects the transformation procedure in \Cref{sec:4.3}.

\noindent\textbf{Pose-mixup Regularization}. Most poses interpolated between two real poses physically exist, and their convex combinations build the real-world pose distribution. Inspired by the {\it mixup} regularization~\cite{zhang2017mixup} on images, we therefore extend it to 2D pose annotations and propose the {\it Pose-mixup} to enrich the guidance set.
The main difference between {\it mixup} and {\it Pose-mixup} is that the {\it mixup} works on {\it image} space and the {\it Pose-mixup} works on {\it keypoint} space. Given that the mixup on skeleton space may lead to unreasonable results,
{\it Pose-mixup} regularizes the neural network to learn the simple linear behavior
in-between 2D poses and thus prevents the model from generating unrealistic human pose annotations. By mixing up the corresponding joints of any two 2D poses with mixup ratio $\lambda\in[0,1]$, the extended annotation set ${T}^*$ from $T$ is then defined as,
\begin{linenomath}
\begin{equation} 
    T^* = \left\{\bm{y}^*~|~\bm{y}^*= \lambda\,\bm{y}_i + (1-\lambda)\,\bm{y}_j,~~~\bm{y}_i,\, \bm{y}_j\in T\right\}.
\end{equation}
\end{linenomath}

\begin{figure*}[tb!]
    \begin{minipage}{.32\linewidth}
        \centering
        \includegraphics[width=0.86\linewidth]{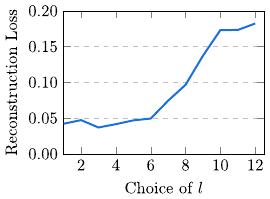} 
        \caption{Reconstruction loss with different choice of layer $l$. }
        \label{fig:layer}
    \end{minipage}
    \hfill
    \begin{minipage}{.32\linewidth}
        \centering
        \includegraphics[width=0.88\linewidth]{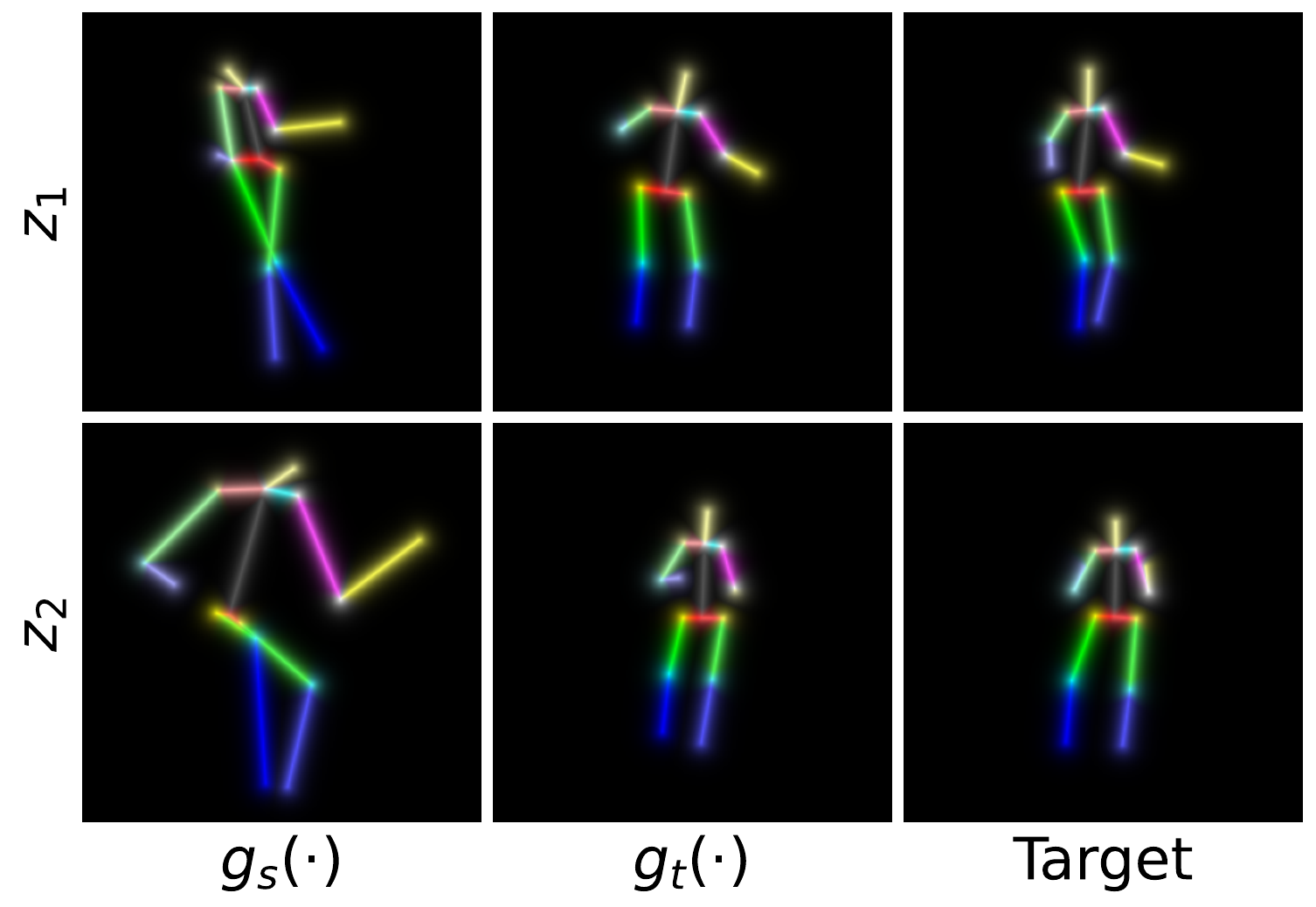}
        \caption{Visualization of pose adaptation. The left and middle of each row are generated from the same random noise. The middle aims to mimic the right.}
        \label{fig:humanpose}
    \end{minipage}
    \hfill
    \begin{minipage}{.32\linewidth}
        \centering
        \includegraphics[width=0.96\linewidth]{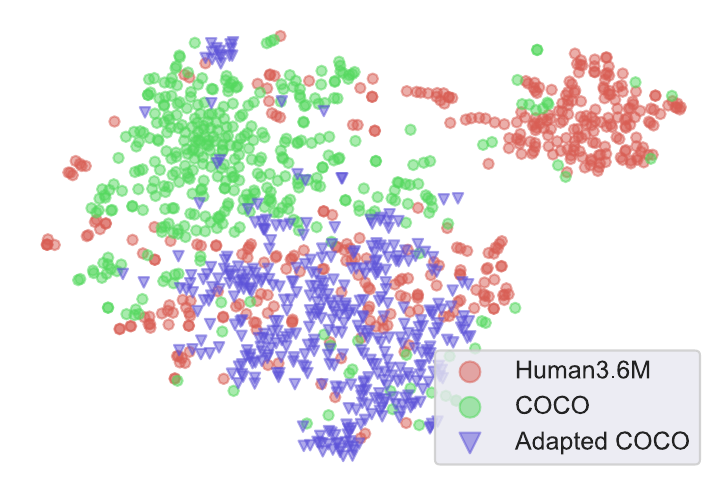} 
        \caption{t-SNE visualization of human poses before and after adaptation. We visualize the pose distribution ($\mathbb{R}^{M\times 2}$) in a two-dimensional space.}
        \label{fig:tsne}
    \end{minipage}
\end{figure*}

\subsection{Target Pose Sampling}
\label{sec:3.4}

Once the transferred generator $g_t(\cdot)$ is obtained, we can generate theoretically as many target skeleton images $\tilde{\bm{x}}_t$ as possible by randomly sampling latent codes in the estimated target distribution ${\mathcal{D}_t}$. Unfortunately, the generated target skeleton images are not perfect and may bring in artifacts which mislead the training of a neural network. 
To address this issue, we utilize $\beta(\cdot)$ to filter out the random noise. $\beta(\cdot)$ is a neural network regressor pre-trained on $S$ and acts as a tight information bottleneck that preserves skeleton information and ignores random noise. Following the generation of the fake skeleton images $\tilde{\bm{x}}_t$ from $g_t(\cdot)$, we extract the coordinates of interpretable 2D keypoints $\hat{T} = \{\hat{{\bm{y}_t}}\}$ from it, {\it e.g.}, hands, by applying $\hat{\bm{y}_t}=\beta(\tilde{\bm{x}}_t)$. Thereafter,
we can get a clean generated skeleton $\hat{\bm{x}}_t$ by a re-render process,
\begin{linenomath}
\begin{equation}
    \hat{\bm{x}}_t = \alpha(\hat{\bm{y}_t}) = \alpha\big(~\beta(\tilde{\bm{x}}_t)~\big).
\end{equation}
\end{linenomath}
These generated skeleton images and the corresponding generated 2D pose annotations can be further utilized to assist any pose-related down-stream tasks.

\begin{table}[tb!]
    \centering
        \begin{tabular}{c|cc|cc}
            \toprule
            Method &Target & Source & MMD$^2$ ($\downarrow$) & FD ($\downarrow$) \\ \midrule
            FreezeD&\multirow{4}*{H3.6M}&COCO& 0.081& 3.77\\
            AdaGAN && COCO & 0.052   & 2.67        \\
            LoRA&&COCO &0.035&1.36\\
            FlexPose& &COCO& \textbf{0.029}&\textbf{0.80}  \\    
            \bottomrule 
        \end{tabular}
    \caption{The results of distribution distance measurement. }
    \label{tab:eva}
\end{table}

\section{Experiments}
\label{sec:4}

In this section, we first evaluate the distribution similarity between transferred distribution and target distribution via two standard metrics. Then, we show how FlexPose can improve the performance of existing unsupervised landmark detection algorithms and benefit unlabelled human pose dataset annotation. At last, we extensively discussed how each part of FlexPose works.

\subsection{Pose Distribution Transformation}
In this subsection, we conduct a transformation experiment between COCO~\cite{lin2014microsoft} and Human3.6M~\cite{h36m_pami} to show how FlexPose works.

\minisection{Experiment Setting}
We train a StyleGAN~\cite{karras2019style} using the skeleton images from the source datasets to estimate the distribution of source human pose. And then we transform the estimated distribution to the target one according to several samples from target dataset.
For source dataset COCO, we only keep the annotated people instances with full pose annotations to construct a training set of 32k samples. The training of StyleGAN follows standard protocol in the original work. 
In the transformation phase, we only use 30 annotations from the target dataset Human3.6M (two for each class). The size of interpolated pose set ($|T^*|$) is set as 1000. We experimented with changing different layers and found that setting $l$=3, \textit{i.e.,} transferring the third coarsest layer, usually gets the lowest reconstruction loss in \Cref{eq:rec} as shown in \Cref{fig:layer}. So, we set $l$=3 in all experiments. A detailed setting and deeper analysis can be found in appendix. When adaptation phase ends, we sample new poses from generator and treat them as Adapted COCO.

\minisection{Evaluation $\&$ Visualization}
Qualitatively, we show the visual result of pose transformation in \Cref{fig:humanpose}. For each row, we show one skeleton (Left) that was randomly sampled from the generator before transformation, one skeleton (Middle) that was sampled from the transformed generator by using the same latent noise as the Left, and one skeleton (Right) in the few-shot annotation set $T$ from target dataset. We can see that the Left and the Middle generated from the same random noise are visually quite different, and the Middle is more similar to the Right.

In \Cref{fig:tsne}, we plot the t-SNE embedding of the poses generated by FlexPose (Adapted COCO), comparing it with the embedding of poses from the source (COCO) and target (Human3.6M) dataset. As can be seen, the embedding of poses from the source and target dataset are separated, and the distribution of generated poses significantly overlaps with the target ones. We also noted that the pose distributions are `mismatched' in the upper right region. Considering that only two shots are utilized as guidance for each class during the transformation, such a mismatch is reasonable.

Quantitatively, we measure the similarity between the transferred distribution and target distribution using the Fréchet distance (\text{\rm FD}), which follows from the Wasserstein-based definition of FID \cite{heusel2017gans} without the application of the pre-trained Inception network. We also measure the square of Maximum Mean Discrepancy (MMD) to provide more insights. The measurements are conducted on the keypoint coordinates space. 

We compare our method with three strong competitors.
AdaGAN~\cite{noguchi2019image}, FreezeD~\cite{mo2020freeze} and LoRA~\cite{hu2021lora} (rank $r$=8). Both AdaGAN and FreezeD suggest finetuning-based strategies with regularization. LoRA is the most related work to our FlexPose, introducing low-rank regularization to the generative model. For all methods, we generate 50k samples from the transferred generative model and compare them with all samples in the target dataset Human3.6M. In \Cref{tab:eva}, experiment results suggest that FlexPose gives superior performance on both MMD and FD evaluation, indicating the transferred distribution shares more similar characteristics to the target distribution. The finding remains the same as that of the observation on qualitative evaluation.

\begin{table}[tb!]
    \centering
    \setlength{\tabcolsep}{1mm}
        \begin{tabular}{c|cc|cc}
            \toprule
            Method &Target & Source & MSE ($\downarrow$) & PCK ($\uparrow$) \\ \midrule
        
            Baseline &\multirow{5}*{H3.6M}& COCO & 17.86   & 0.015        \\
            FreezeD&&COCO&20.60&0.081 \\
            AdaGAN&&COCO&14.88&0.395 \\
            
            LoRA&&COCO&13.85&0.430 \\
            FlexPose& &COCO& \textbf{13.19}&\textbf{0.585}      \\ \midrule
    
            Baseline&\multirow{15}*{S-H3.6M}& COCO & 5.47  &   0.685      \\
            FreezeD&&COCO&7.63&	0.003 \\
            AdaGAN&&COCO&5.36& 0.455 \\
            
            LoRA&&COCO&5.02&0.512 \\
            FlexPose& &COCO&\textbf{3.79}&\textbf{0.770}        \\
            \cline{1-1}
            \cline{4-5}
            Baseline&& 3DHP&12.66& 0.000       \\
            AdaGAN &&3DHP&7.23&0.215 \\
            FreezeD&&3DHP&6.28&	0.206\\
            LoRA&&3DHP&6.15&0.314 \\
            FlexPose&& 3DHP&\textbf{5.98}&\textbf{0.467}        \\
            \cline{1-1}
            \cline{4-5}
            Baseline&& SURREAL&11.18&0.000        \\
            FreezeD&&SURREAL&11.38&	0.006\\
            AdaGAN&&SURREAL&6.63&0.228 \\
            LoRA&&SURREAL&6.52&0.337 \\
            FlexPose&& SURREAL&\textbf{6.47}&\textbf{0.499}        \\
            \bottomrule 
        \end{tabular}

    \caption{Results on human pose annotation task. S-H3.6M and H3.6M are short for Simplified Human3.6M dataset and Human3.6M dataset respectively. The threshold of PCK is 10\% for S-H3.6M and 20\% for H3.6M in this table. }
    \label{tab:simpleh36}
\end{table}

\begin{table}[tb!]
    \centering
        \begin{tabular}{c|cc|cc}
            \toprule
            Method &Target & Source & MSE ($\downarrow$)& PCK ($\uparrow$)\\
            \midrule
            Baseline& \multirow{5}*{WFLW}&300-VW&18.78&0.679\\
            AdaGAN & &300-VW&11.95&\textbf{0.785}\\
            FreezeD& &300-VW&11.66&0.779\\
            LoRA&&300-VW&11.77&0.760 \\
            FlexPose&&300-VW&\textbf{11.64}&0.766 \\
            \bottomrule 
        \end{tabular}

    \caption{Results on human face annotation task. }
    \label{tab:face}
\end{table}

\begin{figure}[tb]
    \centering
    \includegraphics[width=.8\linewidth]{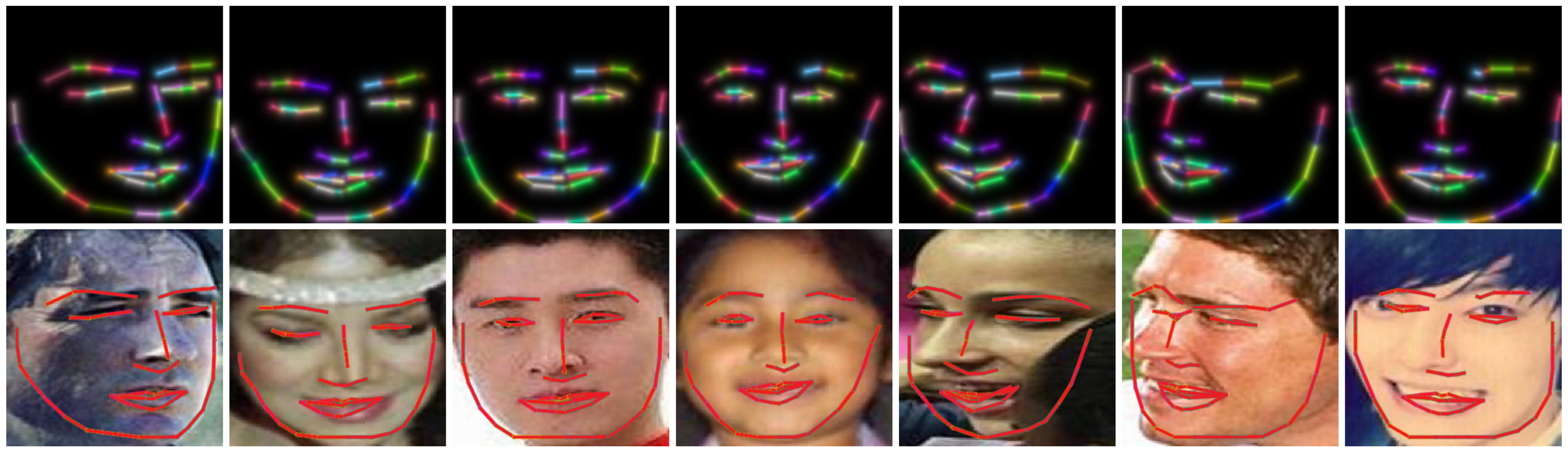}
    \caption{Visualization of human face landmark annotation on WFLW dataset. Upper Row: landmarks given by matching algorithm. Bottom Row: landmarks in the upper row with their corresponding human faces.}
    \label{fig:humanface}
\end{figure}

\subsection{Unlabelled Human Pose Dataset Annotation}

To further evaluate the quality of transformed pose quantitatively and show potential downstream application of FlexPose, we show how can we annotated unlabelled human-related dataset with the help of FlexPose.

\minisection{Pose-Image Matching Algorithm}
In dataset annotation tasks, our goal is to assign each image in the target dataset to the most closely related pose in the estimated target distribution $\mathcal{D}_t$. Existing self-supervised human pose detection methods~\cite{jakab2018unsupervised,lorenz2019unsupervised,thewlis2017unsupervised} are usually constrained by the high relevance between model prediction and input image.
Among them, KeypointGAN~\cite{jakab2020self} can match unpaired images and annotations by forcing the distribution of detector predictions to align with the existing poses. We train KeypointGAN by using human images from target dataset and generated pose set $\hat{T}$. Once the training process is completed, the model prediction on samples can be treated as the best-matched annotations in given distribution.

\minisection{Source Datasets}
Apart from COCO, we also use MPI-INF-3DHP (3DHP)~\cite{mehta2017monocular} which contains more than 1.8 million human pose annotations from eight subjects and covers eight complex exercise activities. SURREAL~\cite{varol17_surreal} is a synthetic dataset containing more than six million frames of people in motion. 

\minisection{Target Datasets} The large-scale dataset \textit{Human3.6M} \cite{h36m_pami} has 3.6 million samples. The \textit{Simplified Human3.6M} dataset \cite{zhang2018unsupervised} contains 800k training and around 90k testing images. We use all human images in the target dataset and randomly select several guide poses.

\minisection{Evaluation Metrics}
Since dataset annotation shares similar targets with 2D landmark detection. 
We report 2D landmark detection performance for evaluation. Two standard evaluation metrics are considered to compare our method with baselines.
The MSE column reports a mean square error in pixels overall pre-defined common joints. The Percentage of Correct Key-points (PCK-$\rho$) is used as an accuracy metric that measures if the distance between the predicted keypoint and the true joint is within a certain threshold $\rho$.

\minisection{Performance Comparisons} 
We feed the generated skeleton images $\hat{\bm{x}}_t$ and RGB human images from target dataset into KeypointGAN to evaluate the effectiveness of each pose transformation algorithm. As a baseline, we train the detector on each target dataset by directly using the pose annotations set $S$ from the source dataset, which we denote as {\it \textbf{Baseline}} in the comparison and can be roughly treated as the worst case. We also employ three strong competitors, AdaGAN, FreezeD, and LoRA, for comparison.

Quantitatively, we compare their performance with FlexPose on human pose estimation in \Cref{tab:simpleh36}. 
As shown in \Cref{tab:simpleh36}, the {\it \textbf{Baseline}} has much lower performance on the target dataset as the pose annotations are from different datasets, especially when some of them have a distinct pose distribution from that of the target dataset, \textit{e.g.,} when 3DHP or SURREAL is the source dataset and Simplified Human3.6M is the target one. FlexPose gets better results on all settings under both metrics. FlexPose largely reduces the performance gap when the pose distribution of the source dataset is very different from that of the target distribution, \textit{e.g.,} MSE 12.7$\rightarrow$6.0 and PCK10 0.00$\rightarrow$0.47 when adaptation occurs from 3DHP to Simplified Human3.6M. The results show that FlexPose is effective at generating similar poses with the target dataset, even with less to only two poses per class in the target dataset.

\subsection{Unlabelled Human Face Dataset Annotation}

Introducing FlexPose to human face landmarks transfer is straightforward since both the human pose and the human face consist of a set of pre-determined keypoints. 

\minisection{Datasets}
\textit{WFLW}~\cite{wayne2018lab} has 10 thousand samples with 98 facial landmarks, where 7.5 thousand for training and 2.5 thousand for testing.
\textit{300-VW}~\cite{sagonas2013300} consists of 300 Videos in the wild and contains $\sim$95 thousand annotated human faces in the training set. 
We treat 300-VW as the source dataset and only use its annotations for training StyleGAN. And few-shot annotations in the target dataset WFLW are utilized for transformation. We only keep the shared 68 facial landmarks in two datasets.

\begin{table}[tb]
    \centering
    \setlength{\tabcolsep}{1mm}
        \begin{tabular}{cccccccc}
            \toprule
            \# & Source & Layer&Mixup&Linear& Shots  & MSE& PCK \\

            \midrule

            1 & \textbf{C}  &3 &$\surd$&$\surd$&12&3.79&0.77\\
            2& \textbf{C}  & 1,3&$\surd$&$\surd$&12&3.82&0.78 \\
            3& \textbf{C}  & 3,5&$\surd$&$\surd$&12&4.02&0.61 \\
            4& \textbf{C}  & ALL&$\surd$&$\surd$&12&4.50&0.66 \\
            \midrule
            5 & \textbf{D}  &3&$\surd$&$\surd$&12&5.98&	0.44\\
            6& \textbf{D}  &ALL&$\times$&$\surd$&12&9.82&	0.01\\
            7& \textbf{D}  &ALL&$\times$&$\times$&12&12.32&0.00\\
            \midrule
            8 & \textbf{C}&3& $\surd$&$\surd$&12&3.79&0.77 \\
            9& \textbf{C}&3& $\surd$&$\surd$&24&3.80&0.75 \\
            10& \textbf{C}&3& $\surd$&$\surd$&48&3.73&0.70\\
            \midrule
            11 & \textbf{D}&3& $\surd$&$\surd$&12&5.98&	0.47 \\
            12& \textbf{D}\textbf{C}&3& $\surd$&$\surd$&12&5.28&0.59\\
            13& \textbf{D}\textbf{C}\textbf{S}&3& $\surd$&$\surd$&12&5.19&0.59 \\
            \bottomrule 
        \end{tabular}

    \caption{
        Ablation study on human pose annotation.
        The target dataset is Simplified-Human3.6M for all experiments. $\textbf{C}$, $\textbf{D}$, $\textbf{S}$ are short for COCO, 3DHP, SURREAL dataset.
    }
    \label{tab:abla}
\end{table}

\minisection{Experiments Settings and Results}
The evaluation metrics and the experiment protocols are the same as that in the human pose. The size of the few-shot guidance set from target dataset is set as 30. We report the evaluation results on the validation set of WFLW in \Cref{tab:face}. FlexPose still outperforms the baseline by a large margin (MSE $18.78\rightarrow11.64$ and PCK $0.679\rightarrow0.766$). Given that the human face can be treated as a rigid body approximately and are easier to transfer, FlexPose achieves comparable performance with previous SOTA methods, AdaGAN, FreezeD and LoRA. 

We show the detected human face landmarks in \Cref{fig:humanface}. The human face detector trained with generated face landmarks can handle human faces in different directions well.

\subsection{Ablation Study \& Parameter Sensitivity}
\label{sec:4.3}

In \Cref{tab:abla}, ablation studies are conducted:

\minisection{Effect of Regularization} We proposed three kinds of regularization in \Cref{sec:3.3} to alleviate the extreme data-insufficient issue. We remove part of them from our FlexPose, and the results are \#1 to \#7. From \#1 to \#4, we gradually relax the sparsity regularization by allowing more blocks in the diagonal matrix $\tau$ not to be an identity matrix $\bm{I}$. The performance only drops by an acceptable level thanks to the Linear and Mixup regularization. Furthermore, in \#5, \#6, and \#7, we further relax the Mixup and Linear regularization, which significantly hurt the quality of generated images and lower the model accuracy in downstream tasks.

\minisection{Number of Shots from Target Dataset} Under the setting of COCO$\rightarrow$S-H3.6M, we increase the number of shots from 12 to 48 and found that the performance of the pose detector has no obvious difference. 
The results can be found in \#8 to \#10. An explanation is that the increment of few-shot samples from the target dataset brings a limited gain of information compared with the strong prior trained on large-scale datasets. Few samples are enough for target distribution localization.

\minisection{Choice of Layers $l$} In previous experiments, we empirically choose $l$=3 in \Cref{eq:h} for all experiments and get significant improvement. We found that the choice of $l$ is not strictly fixed. We have also tried a composition of multi-layer, and the results can be found in \#2, \#3, and \#4. The result in \#4 shows the necessity of sparse regularization. We leave the best choice of $l$ to future work.

\minisection{Multi-source Datasets} To study the effect of the setting where the source annotations are from different datasets, we conduct two additional experiments (\#12 and \#13) in \Cref{tab:abla} and compare them with existing experiments (\#11). In \#12 and \#13, we use the union of different source datasets to train the generic generator. The result indicates that the increasing diversity on the source dataset (\#11$\rightarrow$\#12$\rightarrow$\#13) brings better results on the target dataset. By utilizing FlexPose, the performance of downstream task models can benefit from collecting a more diverse pose dataset, which is much easier compared with collecting a realistic human dataset with accurate landmarks. However, the result of \#13 is still worse than that of \#1, which indicates the trade-off between diversity and similarity to the target dataset when choosing the source.

\section{Conclusion}
\label{sec:6}

We aim to transfer knowledge in the pose domain and propose an effective method named FlexPose. Our approach allows us to adapt an existing pose distribution to a different target one by using a few poses from the target dataset and generating theoretically infinite poses following the target distribution. FlexPose can be used on several pose-related works. In future work, we hope to extend our method to a more generic pose domain adaptation approach.

\clearpage
{
\bibliography{ref/Top, ref/egbib}
}

\newpage
\clearpage
\newpage

\section{Training Details}

\noindent\textbf{Pre-training Setting of StyleGAN.} We follow the settings in StyleGAN~\cite{karras2019style} to learn the source generator. All the hyper-parameters are kept the same, and the generator is trained until convergence. The training set is re-rendered from the pose annotation of the Source Dataset $S$ by the function $\alpha(\cdot)$. Different methods share the common pre-training weights for a fair comparison. 

\noindent\textbf{Training Setting of FlexPose.} We freeze the weights of the pre-trained source generator and only finetune on the linear block-diagonal matrix $\tau(\cdot)$. $\tau(\cdot)$ is initialized with an identity matrix $\bm{I}$, and only a few selected blocks (3rd, for example) of it can be adjusted due to the sparse regularization. The training set only includes the skeletons re-rendered from the target pose set $T$ and the extended annotation set $T^*$. The parameter betas of the Adam optimizer is set as $(0.9,0.999)$. The learning rate is set to 0.1. The batch size is 128 with a training length of 1000 iterations. It takes less than 5 minutes on a single V100 GPU to finetune the model.

\noindent\textbf{Training Setting of Methods for Comparison.} We compare our FlexPose with three similar methods, AdaGAN \cite{noguchi2019image}, FreezeD \cite{mo2020freeze} and LoRA\cite{hu2021lora}, in our experiments. We re-implement AdaGAN under our settings according to the official open-source code. FreezeD has an official implementation on StyleGAN and is utilized for the comparison. Similar to our method, the low-rank modules are conducted aside the linear-block-diagonal matrix $\tau(\cdot)$ with the default rank setting ($r$=8). To have a fair comparison, all the hyper-parameters and the settings including training length, learning rate, batch size, optimizer are well aligned with the competitors.

\section{Implementation of $\alpha(\cdot)$ and $\beta(\cdot)$}

$\alpha(\cdot)$ is a rule-based function. Given an annotation $\bm{y}$, we draw each keypoint on an empty black figure, and then connect them with fixed color lines. The choice of color is random and is pre-defined before the experiment. All methods share the same choice of colors. The visual effect is similar to a stick man.

As a reverse function of $\alpha(\cdot)$, $\beta(\cdot)$ is a pre-trained neural network. The input is a skeleton image $\bm{x}$ and the output is $M$ heatmaps. The locations of keypoints are further obtained by the method introduced in \cite{jakab2018unsupervised} by converting each heatmap into a 2D probability distribution. The training of $\beta(\cdot)$ is offline.  The training procedure is achieved by minimizing the reconstruction loss on the given annotation $\bm{y}$ from the source dataset,

\begin{equation}
    \mathcal{L}_{rec} = \|~\bm{y} - \beta(\alpha(\bm{y}))~\|_2^2.
\end{equation}

\section{Skeleton-guided Applications}

\begin{figure}[tb!]
        \centering
        \includegraphics[width=1.0\linewidth]{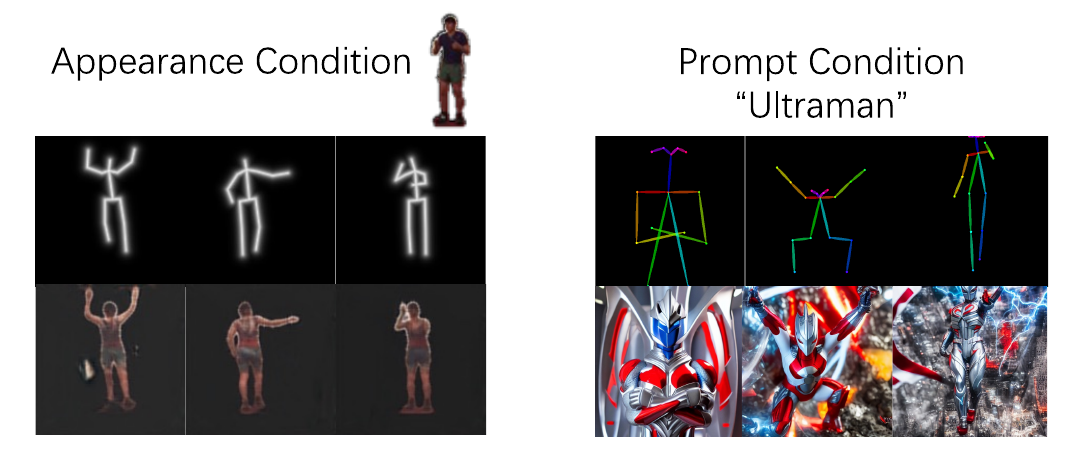}
        \caption{Application on pose-conditional human image generation. FlexPose can synthesize new poses in a certain style, which can be used as conditions for image generation.}
        \label{fig:humangen}
\end{figure}

Recently, there have been studies on generating human images with given 2D poses.
A large amount of reasonable human poses in a certain style or distribution may be needed to evaluate their performance. FlexPose is born for this task and can generate infinite suitable 2D human poses with few-shot human poses in the needed style. 
\Cref{fig:humangen} gives some examples. 
For the appearance-based method, we used the CycleGAN~\cite{zhu2017unpaired} as the generator. For the prompt-based method, we show the result on Stable Diffusion with ControlNet~\cite{zhang2023adding}. {As can be seen, the 2D human poses outputted from FlexPose diversify the human posture in the generated images.}

\begin{figure}[tb!]
    \centering
    \includegraphics[width=0.8\linewidth]{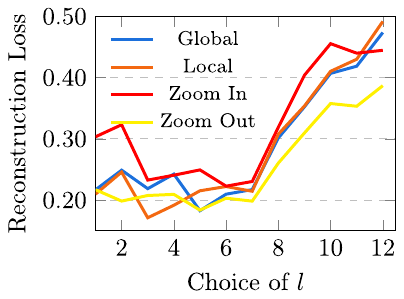}
     \caption{Reconstruction loss with different augmentation methods.}
     \label{appfig:loss}
\end{figure}

\section{Extended Discussion}

We expanded our discussion on the fine-tuned layers selection presented in Figure 4 of the main text. The discussion investigates the pose geometric semantics of each layer in the StyleGAN~\cite{karras2019style}.

\begin{figure*}[th]
        \centering
        \includegraphics[width=.7\linewidth]{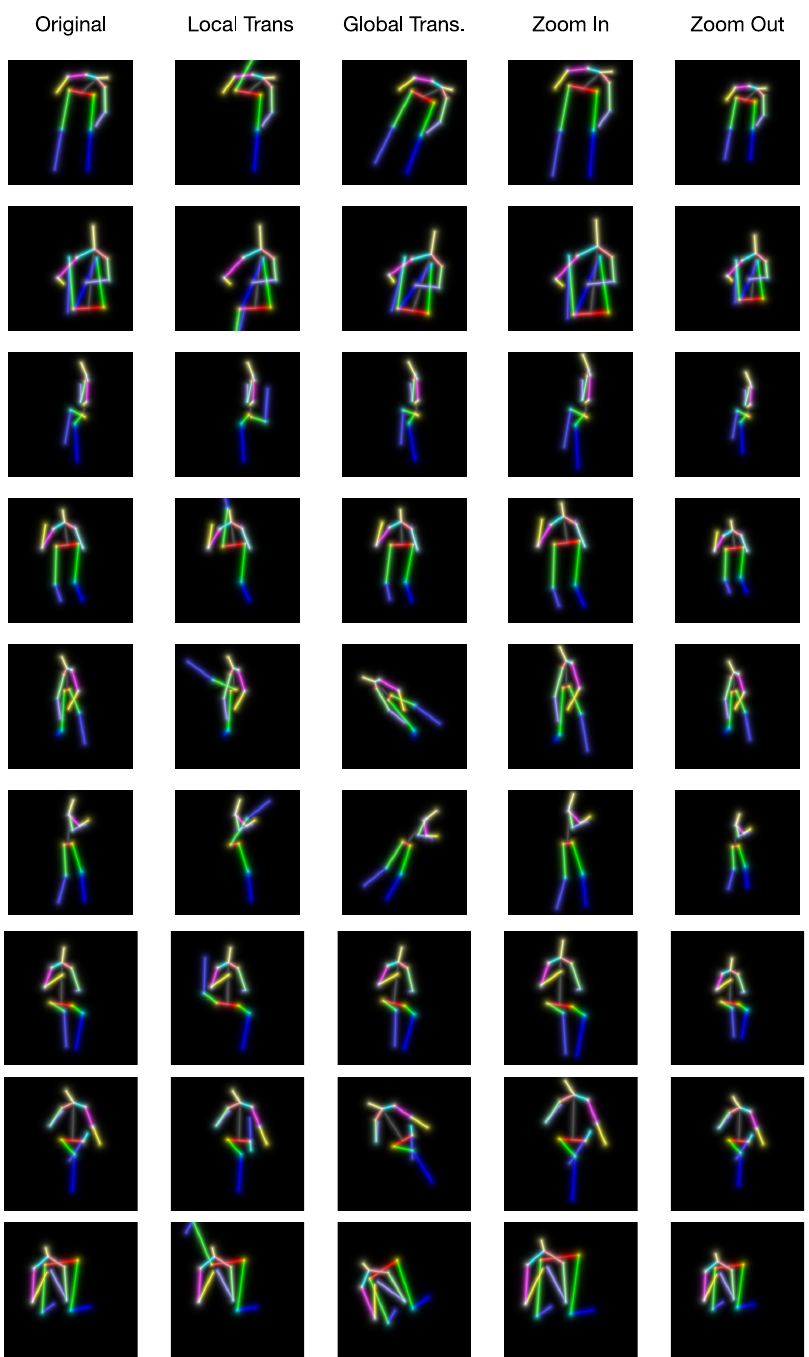}
        \caption{Illustration of augmentations. All of them generate skeleton images that less frequently appear in source targets.}
        \label{appfig:aug}
\end{figure*}

\minisection{Geometric Rotation Transformation.} Two types of {geometric rotation transformation} were applied to the poses from the target dataset. The global {transformation}, $\text{Aug}_G^\theta(\cdot)$, rotates the target skeleton image $\bm{x}$ by a given angle $\theta$. In contrast, the local {transformation}, $\text{Aug}_L^\gamma(\cdot)$, rotates a single leg of the target skeleton image $\bm{x}$ by an angle $\gamma$. The angle $\theta$ is randomly selected from $[-45^\circ, 45^\circ]$, while $\gamma$ is randomly chosen from $[135^\circ, 225^\circ]$. An illustration of these two {transformations} is provided in \Cref{appfig:aug}. The augmented skeleton images can differ significantly from the samples in the source {distribution}. Our investigation centers on identifying which layer {calibration} can minimize the reconstruction loss during the transformation phase when the target skeleton images are individually {transformed} by these two methods. 

\minisection{Scale Transformation.} We also conducted experiments on the scale transformation. We scale the skeleton by $\eta$, where $\eta$ is randomly chosen from $[0.7,0.9]$ and $[1.1,1.2]$. We also showed some examples in \Cref{appfig:aug} and report the reconstruction loss.

The quantitative results are presented in \Cref{appfig:loss}. The findings indicate that coarser layers (layers $3$ and $4$) result in lower reconstruction loss with global augmentation, whereas finer layers (layers $5$ and $6$) are more suitable for local transformations. A noteworthy conclusion is that layers with $l \geq 8$ and $l \leq 2$ have a lesser effect on fitting the distribution of skeleton images. A potential explanation is that the coarsest layers ($l \leq 2$) primarily determine the background, while the finest layers are less associated with skeleton action. This analysis underscores a hierarchical framework in the latent space of skeleton images, extending related insights observed in natural images.

\end{document}